\documentclass[conference, a4paper]{./template/IEEEtran}
\usepackage{cite}

\ifCLASSINFOpdf
  \usepackage[pdftex]{graphicx}
  \graphicspath{{./figures/}}
  \DeclareGraphicsExtensions{.pdf,.jpeg,.png}
\else
  \usepackage[dvips]{graphicx}
  \graphicspath{{./figures/}}
  \DeclareGraphicsExtensions{.eps}
\fi
\usepackage[hidelinks]{hyperref}

\usepackage{pgfplots}
\usepgfplotslibrary{fillbetween,patchplots,groupplots}
\pgfplotsset{compat=newest}

\usepackage{tikz}
\usetikzlibrary{arrows.meta,backgrounds,fit,shapes.geometric,patterns,positioning}

\usepackage{tcolorbox}  

\usepackage{rotating}

\definecolor{rwth}   {RGB}{  0  84 159}
\definecolor{rwth-75}{RGB}{ 64 127 183}
\definecolor{rwth-50}{RGB}{142 186 229}
\definecolor{rwth-25}{RGB}{199 221 242}
\definecolor{rwth-10}{RGB}{232 241 250}

\definecolor{black}   {RGB}{  0   0   0}
\definecolor{black-75}{RGB}{100 101 103}
\definecolor{black-50}{RGB}{156 158 159}
\definecolor{black-25}{RGB}{207 209 210}
\definecolor{black-10}{RGB}{236 237 237}

\definecolor{magenta}   {RGB}{227   0 102}
\definecolor{magenta-75}{RGB}{233  96 136}
\definecolor{magenta-50}{RGB}{241 158 177}
\definecolor{magenta-25}{RGB}{249 210 218}
\definecolor{magenta-10}{RGB}{253 238 240}

\definecolor{yellow}   {RGB}{255 237   0}
\definecolor{yellow-75}{RGB}{255 240  85}
\definecolor{yellow-50}{RGB}{255 245 155}
\definecolor{yellow-25}{RGB}{255 250 209}
\definecolor{yellow-10}{RGB}{255 253 238}

\definecolor{petrol}   {RGB}{  0  97 101}
\definecolor{petrol-75}{RGB}{ 45 127 131}
\definecolor{petrol-50}{RGB}{125 164 167}
\definecolor{petrol-25}{RGB}{191 208 209}
\definecolor{petrol-10}{RGB}{230 236 236}

\definecolor{turkis}   {RGB}{  0 152 161}
\definecolor{turkis-75}{RGB}{  0 177 183}
\definecolor{turkis-50}{RGB}{137 204 207}
\definecolor{turkis-25}{RGB}{202 231 231}
\definecolor{turkis-10}{RGB}{235 246 246}

\definecolor{grun}   {RGB}{ 87 171  39}
\definecolor{grun-75}{RGB}{141 192  96}
\definecolor{grun-50}{RGB}{184 214 152}
\definecolor{grun-25}{RGB}{221 235 206}
\definecolor{grun-10}{RGB}{242 247 236}

\definecolor{maigrun}   {RGB}{189 205   0}
\definecolor{maigrun-75}{RGB}{208 217  92}
\definecolor{maigrun-50}{RGB}{224 230 154}
\definecolor{maigrun-25}{RGB}{240 243 208}
\definecolor{maigrun-10}{RGB}{249 250 237}

\definecolor{orange}   {RGB}{246 168   0}
\definecolor{orange-75}{RGB}{250 190  80}
\definecolor{orange-50}{RGB}{253 212 143}
\definecolor{orange-25}{RGB}{254 234 201}
\definecolor{orange-10}{RGB}{255 247 234}

\definecolor{rot}   {RGB}{204   7  30}
\definecolor{rot-75}{RGB}{216  92  65}
\definecolor{rot-50}{RGB}{230 150 121}
\definecolor{rot-25}{RGB}{243 205 187}
\definecolor{rot-10}{RGB}{250 235 227}

\definecolor{bordeaux}   {RGB}{161  16  53}
\definecolor{bordeaux-75}{RGB}{182  82  86}
\definecolor{bordeaux-50}{RGB}{205 139 135}
\definecolor{bordeaux-25}{RGB}{229 197 192}
\definecolor{bordeaux-10}{RGB}{245 232 229}

\definecolor{violett}   {RGB}{ 97  33  88}
\definecolor{violett-75}{RGB}{131  78 117}
\definecolor{violett-50}{RGB}{168 133 158}
\definecolor{violett-25}{RGB}{210 192 205}
\definecolor{violett-10}{RGB}{237 229 234}

\definecolor{lila}   {RGB}{122 111 172}
\definecolor{lila-75}{RGB}{155 145 193}
\definecolor{lila-50}{RGB}{188 181 215}
\definecolor{lila-25}{RGB}{222 218 235}
\definecolor{lila-10}{RGB}{242 240 247}

\usepackage{amsmath}
\usepackage{amssymb}
\usepackage{bm}  
\usepackage{mathtools}

\DeclarePairedDelimiter{\abs}{\lvert}{\rvert}%
\DeclarePairedDelimiter{\norm}{\lVert}{\rVert}%
\DeclarePairedDelimiterX{\infdivx}[2]{(}{)}{%
	#1\;\delimsize\|\;#2%
}

\usepackage{array}

\usepackage{siunitx}

\usepackage{booktabs}
\usepackage{multirow}

\ifCLASSOPTIONcompsoc
  \usepackage[caption=false,font=normalsize,labelfont=sf,textfont=sf]{subfig}
\else
  \usepackage[caption=false,font=footnotesize]{subfig}
\fi
\usepackage{glossaries}

\usepackage{url}

\begin{document}
%
\title{Modeling the Distribution of Normal Data in Pre-Trained Deep Features for Anomaly Detection}

\author{\IEEEauthorblockN{Oliver Rippel\IEEEauthorrefmark{1}}
\IEEEauthorblockA{\parbox{5.62cm}{\noindent Institute of Imaging \& Computer Vision}\\
RWTH Aachen University\\
Aachen, Germany\\
Email: oliver.rippel@lfb.rwth-aachen.de}
\and
\IEEEauthorblockN{Patrick Mertens\IEEEauthorrefmark{1}}
\IEEEauthorblockA{\parbox{5.62cm}{\noindent Institute of Imaging \& Computer Vision}\\
RWTH Aachen University\\
Aachen, Germany}
\and
\IEEEauthorblockN{Dorit Merhof}
\IEEEauthorblockA{\parbox{5.62cm}{\noindent Institute of Imaging \& Computer Vision}\\
RWTH Aachen University\\
Aachen, Germany}
}

\hypersetup{
	pdftitle    = {Modeling the Distribution of Normal Data in Pre-Trained Deep Features for Anomaly Detection},
	pdfsubject  = {Anomaly Detection},
	pdfauthor   = {Oliver Rippel*, Patrick Mertens*, Dorit Merhof},
	pdfkeywords = {Anomaly Detection, Deep Learning, MVTec AD, Mahalanobis, Autoencoder}
}


%


\maketitle
\begingroup\renewcommand\thefootnote{\IEEEauthorrefmark{1}}
\footnotetext{Authors contributed equally to this work}
\endgroup
\begin{abstract}
Anomaly Detection (AD) in images is a fundamental computer vision problem and refers to identifying images and/or image substructures that deviate significantly from the norm.
Popular AD algorithms commonly try to learn a model of normality from scratch using task specific datasets, but are limited to semi-supervised approaches employing mostly normal data due to the inaccessibility of anomalies on a large scale combined with the ambiguous nature of anomaly appearance.

We follow an alternative approach and demonstrate that deep feature representations learned by discriminative models on large natural image datasets are well suited to describe normality and detect even subtle anomalies in a transfer learning setting.
Our model of normality is established by fitting a multivariate Gaussian (MVG) to deep feature representations of classification networks trained on ImageNet using normal data only.
By subsequently applying the Mahalanobis distance as the anomaly score we outperform the current state of the art on the public MVTec AD dataset, achieving an Area Under the Receiver Operating Characteristic curve of $\mathbf{95.8 \pm 1.2 \%}$ (mean $\pm$ SEM) over all 15 classes.
We further investigate why the learned representations are discriminative to the AD task using Principal Component Analysis.
We find that the principal components containing little variance in normal data are the ones crucial for discriminating between normal and anomalous instances.
This gives a possible explanation to the often sub-par performance of AD approaches trained from scratch using normal data only.
By selectively fitting a MVG to these most relevant components only, we are able to further reduce model complexity while retaining AD performance.
We also investigate setting the working point by selecting acceptable False Positive Rate thresholds based on the MVG assumption.
Code is publicly available at \url{https://github.com/ORippler/gaussian-ad-mvtec}. 
\end{abstract}


%
\IEEEpeerreviewmaketitle

\section{Introduction}
Anomaly Detection (AD) relates to identifying instances in data that are significantly different to the norm \cite{Chandola2009,Pimentel2014}.
Correspondingly, AD in images aims at finding irregularities in images and poses a fundamental computer vision problem with various application domains ranging from industrial quality control \cite{Bergmann2019} to medical image analysis \cite{Schlegl2019}. 
In general, AD tasks are defined by the following two characteristics:
\begin{itemize}
  \item Anomalies are rare events, i.e.\ their prevalence in the application domain is low.
  \item Anomaly appearance is not well-defined (i.e.\ anomalies types are ambiguous).
\end{itemize}
Together, these characteristics result in AD datasets that are heavily imbalanced, often containing only few anomalies for model verification and testing.

As a consequence, AD algorithms often focus on semi-supervised learning approaches, where a model of normality is established based on normal data only \cite{Bergmann2019, Schlegl2019, Ruff2018}.
While small dataset sizes predestine the capitalization of pre-training on large-scale databases such as ImageNet \cite{Deng2009}, only little research is performed to explore this potential \cite{Napoletano2018, Andrews2016, Cohen2020}.
Instead, methods focus on learning feature representations from scratch, often in reconstruction-based approaches~\cite{Gong2019, Schlegl2019}.

As our main contribution, we demonstrate the effectiveness of pre-trained deep feature representations transferred to the AD task. 
By fitting a multivariate Gaussian (MVG) to normal data of deep features learned by ImageNet training and using the Mahalanobis distance \cite{Mahalanobis1936} as the anomaly score, we are able to outperform the prior state of the art on the public MVTec AD dataset \cite{Bergmann2019}.
We additionally gain insight into and explain the discriminative nature of pre-trained deep features by means of Principal Component Analysis (PCA).
Here, we find that principal components that retain little variance in normal data are highly discriminative to the AD task, indicating that learning these features from scratch may be difficult using normal data only.
We further show that the working point can be sensibly set based on choosing an acceptable False Positive Rate (FPR) under the MVG assumption. 
Here, retaining only highly variant principal components decreases FPR at the cost of AD performance. 
These results demonstrate that there should be a clear focus on leveraging pre-trained deep feature representations in future AD research.

\section{Related Work}
In recent years, a large body of research has been published in the field of AD.
Therefore we provide an extensive overview of AD techniques in the following, focussing on methods applied to image data.
We further categorize the approaches into whether they leverage pre-trained deep feature representations in a transfer learning approach or are learned from scratch.

\subsection{Learning AD from Scratch}
Learning useful representations from scratch in a semi- or unsupervised manner is a major research field on its own. 
Out of the multitude of ways of learning such representations, autoencoder-based approaches are the most popular in AD.

Here, autoencoders (AEs) try to learn the identity function in a semi-supervised manner using a given set of exclusively normal training images. 
The learned identity function is constrained, whereas the model first has to compress the input image to a low dimensional embedding, and subsequently has to reconstruct the input image based on this embedding.
It is argued that the overall model cannot represent anomalous image structures, reconstructing a plausible normal image instead.
An AE trained until convergence can then be used for AD in different ways:

Anomalous images can be detected by comparing the input test image with its reconstruction yielded by the model.
There have been various proposals employing this reconstruction for AD \cite{Bergmann2018, Haselmann2018, Gong2019}.
While the results of reconstruction-based AD approaches are intuitive to understand, they suffer from two drawbacks:
(I) The reconstruction has to be post-processed in order to yield an image-level anomaly score, thus increasing the complexity of the method and (II) the decoder part introduces additional computational overhead.

Embeddings learned by AEs are also utilized in many AD frameworks.
Common approaches try to model the distribution of normal data in the AE embedding in a generative way using variational AEs \cite{Daniel2019} that are oftentimes trained using an adversarial objective \cite{Pidhorskyi2018}.
Alternatively, classical shallow ML methods such as $k$-Nearest Neighbor ($k$-NN) or one class Support Vector Machine (oc-SVM) \cite{Schoelkopf2001} are also applied to embeddings learned by an AE \cite{Sarafijanovic-Djukic2019}.
More recently, Ruff et al.\ have initialized their proposed Deep Support Vector Data Description (Deep SVDD) using pre-trained AEs \cite{Ruff2018, Ruff2020}.

Hybrid approaches also exist, where anomaly scores are generated by combining measures proposed on the embeddings with reconstruction errors \cite{Abati2019, Zong2018, Vasilev2018}, enhancing model performance at cost of further increased complexity.

\subsection{Transfer Learning AD with Deep Feature Representations}
Less extensively studied than semi-supervised feature learning methods, AD has also been performed by using deep representations learned by large-scale ImageNet training in a transfer learning setting for both anomaly segmentation and image-level AD.

While there has been recent success in adapting deep feature representations for anomaly segmentation \cite{Sabokrou2018, Napoletano2018, Bergmann2020}, these proposals compute features patch-wise to yield the pixel-wise output. 
As a consequence, receptive fields are limited, feature complexity is rather low and there is an implicit assumption that the anomalies fit inside one patch.
Further, segmentations have to be aggregated to yield image-level AD.
Regarding image-level AD, Christiansen et al.\ \cite{Christiansen2016} repurpose deep AlexNet \cite{Krizhevsky2012} and VGG \cite{Simonyan2015} features for agricultural anomalous object detection.
While they also fit a MVG to deep feature representations and use the Mahalanobis distance as an anomaly measure, they evaluate their model using a small in-house dataset only.
Further, in their use-case anomalous instances deviate significantly in appearance from the normal class (as opposed to the subtle deviations present in the MVTec AD dataset), and benchmarking against other AD approaches is not performed. 
Also, they do not investigate the properties of the pre-trained feature representations that make them suitable to AD.
Andrews et al.\ successfully fit an oc-SVM to deep representations learned by VGG on ImageNet for AD in X-Ray scans of containers \cite{Andrews2016}.
Bergman et al.\ \cite{Bergman2020a} and Cohen et al.\ \cite{Cohen2020} evaluate a $k$-NN using L2-distance on ResNet \cite{He2016} features.
Here, Cohen et al.\ \cite{Cohen2020} report an average Area Under the Receiver Operating Characteristic curve (AUROC) of 85.5\% on the public MVTec AD dataset with $k=50$ and average-pooled features extracted from the last convolutional layer of a Wide-ResNet50-2.
Except for these and a 1-NN approach with different normalizations in surveillance videos \cite{Nazare2018}, little notice has been given to employing deep features in AD for classifying full images. 

While not directly used as an AD algorithm, Lee et al.\ also model the data distribution of in-distribution data by means of MVG for Out-Of-Distribution (OOD) detection \cite{Lee2018}.
Contrary to AD, OOD determines whether a given query image is part of the in-distribution dataset (i.e.\ the dataset used for training) or OOD.
Using a small subset of anomalies for fine-tuning, they apply the linear combination of Mahalanobis distances computed at various depths of a pre-trained ResNet \cite{He2016} to a test image.
The test image is additionally pre-processed to maximize Mahalanobis distance by means of performing a single gradient ascent step to implicitly evaluate Probability Density Function (PDF) around the test image. 
They further show that discriminative deep classifiers employing softmax learn the same posterior distribution as generative classifiers under a Linear Discriminant Analysis (LDA) assumption.
They expand on this and argue that the pre-trained features of the deep softmax classifier may also follow the class-conditional Gaussian distribution of the generative classifier. 
While they do not give a theoretical proof for this, the performance of a generative classifier based on pre-trained features is verified experimentally.
They also show that prior unseen classes may be easily integrated into the generative classifier by introducing a new class to the LDA (compute new mean and update joint covariance).
This finding is the motivation for our work, where we apply pre-trained deep feature representations to the AD task in a transfer learning setting.

\section{Modeling Normal Data Distribution in Deep Feature Representations}
Based on the findings of Lee et al.\ \cite{Lee2018}, we hypothesize that pre-trained deep representations can also be successfully applied to the AD task.
Similar to the class-incremental learning approach, we directly model the PDF of each "normal" class in the pre-trained features, using normal data only and omitting fine-tuning of the pre-trained model.

We model the PDF using the MVG, defined as
\begin{equation}
 	\varphi_{\bm{\mu},\Sigma}(\mathbf{x}) := \frac{1}{\sqrt{(2\pi)^D \abs{\det\Sigma}}} e^{-\frac{1}{2} (\mathbf{x} - \bm{\mu})^\top \Sigma^{-1} (\mathbf{x} - \bm{\mu})}.
\end{equation}
Here, $D$ is the number of dimensions, $\bm{\mu} \in \mathbb{R}^D$ is the mean vector and $\Sigma \in \mathbb{R}^{D\times D}$ the symmetric covariance matrix of the distribution.
$\Sigma$ must be positive definite.

Under a Gaussian distribution with mean $\bm{\mu}$ and covariance $\Sigma$, a distance measure between a particular point $\mathbf{x} \in \mathbb{R}^D$ and the distribution is called the Mahalanobis distance and defined as
\begin{equation}
	M(\mathbf{x}) = \sqrt{\left(\mathbf{x} - \bm{\mu}\right)^\top \Sigma^{-1} \left(\mathbf{x} - \bm{\mu}\right)}.
\end{equation}
Introduced by Mahalanobis in 1936, $M(\mathbf{x})$ is a useful measure of a sample's uncertainty \cite{Mahalanobis1936}.
This interpretation stems from the fact that the Mahalanobis distance uniquely determines the probability density $\varphi_{\bm{\mu},\Sigma}(\mathbf{x})$ of an observation.
When $\mathbf{x}$ is sampled from the Gaussian distribution, $M(\mathbf{x})^2$ is chi-squared distributed with $k=D$ degrees of freedom.
This $\chi^2$-distribution with $k$ degrees of freedom is the sum of $k$ conditionally independent standard normal random variables.
Its PDF is given as
\begin{equation}
	f_k(x) = \begin{cases}
		{\dfrac {x^{{\frac {k}{2}}-1}e^{-{\frac {x}{2}}}}{2^{\frac {k}{2}}\Gamma \left({\frac {k}{2}}\right)}}, & x>0; \\
		0, & {\text{otherwise}}.
	\end{cases}
\end{equation}
Here, $\Gamma(s)$ is the gamma function for $s > 0$.
The Cumulative Distribution Function (CDF) of the $\chi^2$-distribution is calculated as
\begin{equation}
	F_k(x) = \frac{\gamma({\frac{k}{2}},\,{\frac{x}{2}})}{\Gamma({\frac{k}{2}})}
\end{equation}
with the lower incomplete gamma function $\gamma(s,x)$.

\subsection{Covariance Estimation}

As the true distribution of the novel data in the deep feature spaces is unknown, the covariance matrix $\Sigma$ needs to be approximated from observations $\mathbf{x}_1,\dots,\mathbf{x}_n \in \mathbb{R}^D$ with the sample covariance 
\begin{equation}
	\hat{\Sigma} = \frac{1}{n-1} \sum_{i=1}^{n}\left(\mathbf{x}_i - \mathbf{\bar{x}}\right) \left(\mathbf{x}_i - \mathbf{\bar{x}} \right)^\top.
\end{equation}
Here, $\mathbf{\bar{x}}$ denotes the empirical mean of the observations.

However, the sample covariance matrix is only well-conditioned when the number of dimensions $D$ is much lower than the number of samples $n$.
If $\frac{D}{n}$ is non-negligible, the covariance estimate becomes unstable, and when $D > n$, $\hat{\Sigma}$ becomes singular and hence is not invertible. 
To solve this problem the concept of shrinkage has been proposed for sample covariance estimation and will be used to estimate the sample covariance in this work.

Shrinkage is defined as a linear combination of empirically estimated $\hat{\Sigma}$ and the (scaled) identity matrix $I_D$,
\begin{equation}
	\hat{\Sigma}_\text{shrunk} = (1-\rho)\hat{\Sigma} + \rho \frac{\operatorname{tr}(\hat{\Sigma})}{D} I_D
\end{equation}
with shrinkage intensity $\rho$.
Thus, $\rho$ regulates the influence of the empirical estimator on the final matrix, preferring the well-conditioned identity matrix for larger $\rho$.
It can be seen as a bias-variance tradeoff between the biased, invariant identity estimate and the unbiased high-variance empirical covariance.

By minimizing the expected squared error $\operatorname{E}[\norm{\hat{\Sigma}_\text{shrunk} - \Sigma}^2]$ to the true covariance, Ledoit, Wolf et al.\ obtain a closed form solution for the amount of shrinkage that allows optimal selection of $\rho$ given an unstable estimate of $\hat{\Sigma}$ \cite{Ledoit2004}.

\subsection{Setting the Working Point}
In the case of assuming an underlying MVG, the working point can be estimated based on probabilities.
The idea is that if a specific Mahalanobis distance corresponds to a probability $p$ of seeing a normal sample, this matches the expected True Negative Rate (TNR) of a detector thresholded at that distance.
$1-p$ can be seen as the allowed probability of falsely-labeled normal instances, i.e.\ the FPR.

For a (multivariate) Gaussian the probability of seeing a sample with a Mahalanobis score less than $t$ with $t > 0$ is given by the CDF $F_D$ of the chi-square distribution as
\begin{equation}
	1-\text{FPR} = P(M < t) = P(M^2 < t^2) = F_D(t^2) = \frac{\gamma\mathopen{}\left(\frac{D}{2}, \frac{t^2}{2}\right)\mathclose{}}{\Gamma\mathopen{}\left(\frac{D}{2}\right)\mathclose{}}.
\end{equation} 

Solving for $t$, the AD threshold is obtainable using the inverse CDF for any desired FPR.
\begin{equation} \label{eq:threshold_fpr}
	t = \sqrt{F_D^{-1}(1-\text{FPR})}.
\end{equation}

\section{Experiments and Results}

\begin{figure}
	\centering
	\begin{tikzpicture}[x=0.47cm,
	layer/.style={trapezium, trapezium stretches body, trapezium angle=61.99, shape border rotate=270, draw, minimum height=1pt},
	rect/.style={trapezium, trapezium stretches body, trapezium angle=90, shape border rotate=270, draw, minimum height=1pt},
	attach/.style={below, align=center},
	arr/.style={{Circle[red,length=4pt]}-Latex,shorten <=-2pt},
	hw/.style={rotate=90,above,align=center},
	layer2/.style={layer, path picture={
			\draw (path picture bounding box.north) -- (path picture bounding box.south);
	}},
	layer3/.style={layer, path picture={
			\draw ($(path picture bounding box.north west)!.6666!(path picture bounding box.north)$) -- ($(path picture bounding box.south west)!.6666!(path picture bounding box.south)$);
			\draw ($(path picture bounding box.north)!.3333!(path picture bounding box.north east)$) -- ($(path picture bounding box.south)!.3333!(path picture bounding box.south east)$);
	}},
	rect3/.style={rect, path picture={
			\draw ($(path picture bounding box.north west)!.6666!(path picture bounding box.north)$) -- ($(path picture bounding box.south west)!.6666!(path picture bounding box.south)$);
			\draw ($(path picture bounding box.north)!.3333!(path picture bounding box.north east)$) -- ($(path picture bounding box.south)!.3333!(path picture bounding box.south east)$);
	}},
	layer4/.style={layer, path picture={
			\draw ($(path picture bounding box.north west)!.5!(path picture bounding box.north)$) -- ($(path picture bounding box.south west)!.5!(path picture bounding box.south)$);
			\draw (path picture bounding box.north) -- (path picture bounding box.south);
			\draw (path picture bounding box.north west) -- (path picture bounding box.south west);
			\draw ($(path picture bounding box.north)!.5!(path picture bounding box.north east)$) -- ($(path picture bounding box.south)!.5!(path picture bounding box.south east)$);
	}}
	]
	\node[layer, minimum width=5cm] (block0) at (0,0) {};
	\node[rect, minimum width=4.5cm] (block1) at (2,0) {};
	\node[layer2, minimum width=4.0cm] (block2) at (4,0) {};
	\node[layer2, minimum width=3.0cm] (block3) at (6,0) {};
	\node[layer3, minimum width=2.0cm] (block4) at (8,0) {};
	\node[rect3, minimum width=1.5cm] (block5) at (10,0) {};
	\node[layer4, minimum width=1.0cm] (block6) at (12,0) {};
	\node[rect, minimum width=0.5cm] (block7) at (14,0) {};
	\node[rect, minimum width=0.5cm] (block8) at (16,0) {};
	
	\node[hw,yshift=2pt](input) at (0,0) {$224 \times 224$};
	\draw[arr] (1.3,0) node[hw]{$112 \times 112$} -- (1.3,-3)  node[attach]{$32$};
	\draw[arr] (3.3,0) node[hw]{$112 \times 112$} -- (3.3,-3)  node[attach]{$16$};
	\draw[arr] (5.3,0) node[hw]{$56 \times 56$} -- (5.3,-3)  node[attach]{$24$};
	\draw[arr] (7.3,0) node[hw]{$28 \times 28$} -- (7.3,-3)  node[attach]{$40$};
	\draw[arr] (9.3,0) node[hw]{$14 \times 14$} -- (9.3,-3)  node[attach]{$80$};
	\draw[arr] (11.3,0) node[hw]{$14 \times 14$} -- (11.3,-3)  node[attach]{$112$};
	\draw[arr] (13.3,0) node[hw]{$7 \times 7$} -- (13.3,-3)  node[attach]{$192$};
	\draw[arr] (15.3,0) node[hw]{$7 \times 7$} -- (15.3,-3)  node[attach]{$320$};
	\draw[arr] (17.3,0) node[hw]{$7 \times 7$} -- (17.3,-3)  node[attach]{$1280$};

	\foreach \x [count=\xi from 1] in {1.3, 3.3, 5.3, ..., 17.3}
	{\draw[shorten <=-2pt, -Latex] (\x, -3.6) -- (\x, -4) node[attach]{$M_{\xi}$};}
	\node[align=center,anchor=south] (index4) at (9,-6.0) {$\displaystyle \sum_{i=1}^{9} \limits M_i$};
	
	\node[anchor=south] (index0) at (1,2) {1};
	\node[anchor=south] (index1) at (3,2) {2};
	\node[anchor=south] (index2) at (5,2) {3};
	\node[anchor=south] (index3) at (7,2) {4};
	\node[align=center,anchor=south] (index4) at (9,2) {Level\\5};
	\node[anchor=south] (index5) at (11,2) {6};
	\node[anchor=south] (index6) at (13,2) {7};
	\node[anchor=south] (index7) at (15,2) {8};
	\node[anchor=south] (index8) at (17,2) {9};
	\end{tikzpicture}
	\caption{Anomaly Detection using pre-trained deep feature representations.
	Fitting a multivariate Gaussian to features extracted from every level of an ImageNet pre-trained model and subsequently applying the Mahalanobis distance as anomaly score followed by their unweighted summation yields a simple yet effective Anomaly Detection algorithm.
	The figure depicts this procedure for the EfficientNet-B0 architecture.}
	\label{fig:features_efficientnet}
\end{figure}
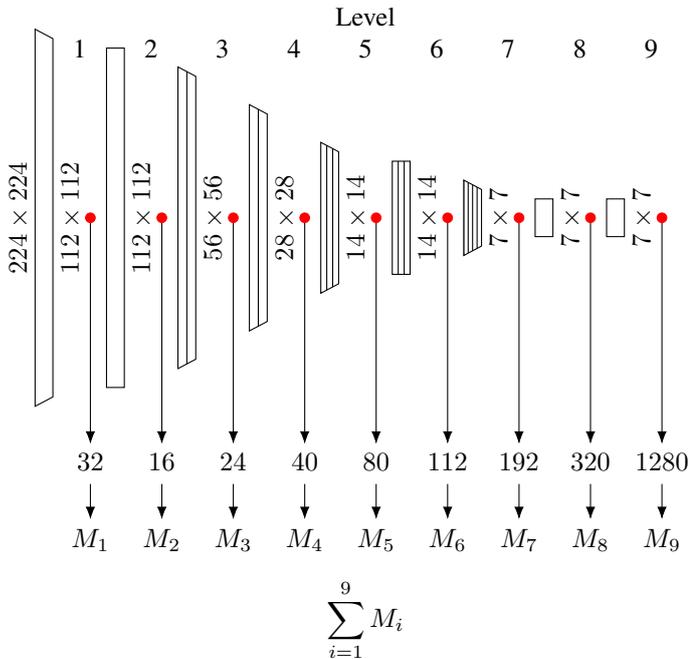

First, we assess the suitability of deep features extracted at various stages of a pre-trained classifier model for AD. 
We employ EfficientNet, which achieves state-of-the-art accuracy on ImageNet classification \cite{Tan2019}, as well as ResNet \cite{He2016}, a commonly applied model in research, as architecture variants.
We extract features at the end of every model block ``level" to assess which feature level gives the best performance.
Here, ``level" is defined as in \cite{Tan2019,He2016} (cf. Appendix Table~\ref{tab:efficient-b0_baseline} for EfficientNet-B0). 
We argue that class probabilities are too application-specific and therefore make use of the features before the final mapping in the highest level.
As feature probability maps may contain spatial dimensions in earlier levels, aggregation is necessary.
We choose simple average pooling to reduce the complexity of our approach, but it should be noted that dedicated aggregation procedures may be an avenue of future research, especially for smaller anomalies.
To guarantee reproducibility of our work, we make our code publicly available\footnote{\url{https://github.com/ORippler/gaussian-ad-mvtec}}.
The overall approach is depicted in Fig.~\ref{fig:features_efficientnet}.

We now compare our approach to two different assumptions:
(I) When assuming a fixed-variance univariate Gaussian distribution, the anomaly score reduces to the simple $L_2$-distance to the mean of the training set. 
(II) When assuming a feature-independent univariate Gaussian, an anomaly score can be defined with the \emph{standardized Euclidean distance (SED)}: 
\begin{equation}
	S(\mathbf{x}) := \sqrt{\sum_{d=0}^{D} \frac{\left(f_d(\mathbf{x}) - \mathbf{\bar{f}}_d\, \right)^2}{s_d^2}}.
\end{equation}
Here, $s_d$ is the (empirical) standard deviation of the $d$-th feature in the training set, and $\mathbf{\bar{f}}_d$ its empirical mean.

To increase robustness of our evaluation, we perform a 5-fold evaluation over the original training dataset of each MVTec category, where we compute the necessary characteristics for each fold, respectively, and apply the scores to the test set of MVTec AD.
We refer to the original publication \cite{Bergmann2019} for an overview of the MVTec AD dataset and to Appendix Figure~\ref{fig:representative_anomalies} for additional, representative anomalies.
To fully assess the capability of the approach, we compute and compare the AUROC, a commonly employed measure for binary classification problems \cite{Ferri2011}.
We report mean $\pm$ Standard Error of the Mean (SEM) AUROC performance over all categories and folds in percent.
In addition to the feature level performances, we also report the AUROC performance yielded by summing distance scores over all levels.
Further, note that no feature reduction is performed in this first evaluation.

\begin{table}
  \caption{Feature level AUROC ($\pm$ SEM) scores in percent for EfficientNet-B4 using different normal distributions.}
	\label{tab:features_comparison_efficientnet}
	\centering
	\begin{tabular}{@{}crrrrrr@{}}
		\toprule
		\multirow{2}{*}{Level} & \multicolumn{2}{c}{$L_2$} & \multicolumn{2}{c}{SED} & \multicolumn{2}{c@{}}{Mahalanobis} \\
		\cmidrule(lr){2-3}
		\cmidrule(lr){4-5}
		\cmidrule(l){6-7}
		& Mean & SEM & Mean & SEM & Mean & SEM \\
		\midrule
		1 & 44.5 & 4.8 & 51.6 & 5.7 & 60.3 & 6.1 \\
		2 & 47.3 & 5.3 & 48.1 & 5.1 & 62.0 & 6.4 \\
		3 & 58.1 & 5.9 & 59.2 & 6.3 & 71.1 & 5.4 \\
		4 & 59.7 & 4.7 & 61.5 & 5.1 & 75.6 & 5.5 \\
		5 & 62.6 & 4.8 & 66.1 & 5.0 & 82.1 & 4.6 \\
		6 & 71.7 & 4.4 & 74.3 & 4.3 & 89.1 & 3.1 \\
		7 & 82.9 & 4.3 & 85.1 & 4.0 & 96.7 & 1.0 \\
		8 & 83.2 & 3.7 & 85.2 & 3.4 & 95.5 & 1.1 \\
		9 & 83.3 & 3.7 & 87.8 & 3.0 & 93.1 & 1.7 \\
		\midrule
		Sum & 75.3 & 4.5 & 79.5 & 6.6 & 94.8 & 1.6 \\
		\bottomrule
	\end{tabular}
\end{table}

Assessing the performance of the three different normal distributions with their respected anomaly scores for EfficientNet-B4 in Table~\ref{tab:features_comparison_efficientnet}, the following two observations can be made:
(I) The MVG is best suited for AD due to its high and robust performance (with an AUROC of 96.7\%~$\pm$~1.0\% for level 7).
(II) Deeper feature representations are more suitable for AD in a transfer learning setting.
This is congruent with findings reported by \cite{Andrews2016}, and reasons for this may be found in the increased abstraction level that is necessary to conclusively describe the distribution of normality. 
However, performance saturates (and even starts to decline) in higher levels (level 8 and 9) in case of the multivariate approach.

Comparing model architectures, features extracted from ResNet models yield worse performance as indicated by the lower average AUROC of $90.4\% \pm 3.6\%$ for the best level~4 and $88.2\% \pm 4.0\%$ for the sum predictor in ResNet-34 (cf. Appendix Table~\ref{tab:features_resnet_maha}).
The increased AD performance of EfficientNet may be attibuted to its efficient architecture (i.e.\ higher ImageNet accuracy per trainable weight) and the output range of the Swish activation function \cite{Ramachandran2017}.
In fact, SED score calculation with features extracted after the ReLU activation used in ResNet often failed as the activations for normal data are clipped to zero for some features.

Compared to OOD \cite{Lee2018}, no learned, linearly weighted sum of feature-level anomaly score is required to achieve strong performance. 
In fact, average AUROC of 94.8\% $\pm$ 1.6\% is achieved by simple equal weighting.
While Hsu et al.\ \cite{Hsu2020} show that this linear weighting of feature level distributions is also not strictly necessary for OOD, their OOD approach still relies on input preprocessing by means of gradient ascent.
It should also be noted that OOD, although similar to AD, still ultimately pursues a slightly different objective.


We also evaluate the influence of model complexity on AD performance of deep features in a transfer learning setting and apply the proposed method to all EfficientNet variants.

\begin{table*}
	\caption[AUROC ($\pm$ SEM) scores in percent for EfficientNet features with Mahalanobis distance]{AUROC ($\pm$ SEM) scores in percent for EfficientNet features with Mahalanobis distance}
	\label{tab:features_mahalanobis_efficientnet}
	\centering
	\begin{tabular}{@{}crrrrrrrrrrrrrrrr@{}}
		\toprule
		\multirow{2}{*}{Level} & \multicolumn{2}{c}{EfficientNet-B0} & \multicolumn{2}{c}{EfficientNet-B1} & \multicolumn{2}{c}{EfficientNet-B2} & \multicolumn{2}{c}{EfficientNet-B3} & \multicolumn{2}{c}{EfficientNet-B4} & \multicolumn{2}{c}{EfficientNet-B5} & \multicolumn{2}{c}{EfficientNet-B6} & \multicolumn{2}{c@{}}{EfficientNet-B7} \\
		\cmidrule(lr){2-3}
		\cmidrule(lr){4-5}
    	\cmidrule(lr){6-7}
    	\cmidrule(lr){8-9}
    	\cmidrule(lr){10-11}
		\cmidrule(lr){12-13}
		\cmidrule(lr){14-15}
		\cmidrule(l){16-17}
		  & Mean & SEM & Mean & SEM & Mean & SEM & Mean & SEM & Mean & SEM & Mean & SEM & Mean & SEM & Mean & SEM \\
		\midrule
		1   & 56.8 & 6.0 & 56.7 & 6.0 & 59.9 & 6.1 & 60.1 & 6.3 & 60.3 & 6.1 & 61.5 & 6.2 & 61.3 & 6.3 & 60.4 & 6.2 \\
		2   & 62.3 & 5.7 & 58.2 & 6.0 & 59.5 & 5.7 & 62.0 & 6.2 & 62.0 & 6.4 & 63.7 & 6.5 & 63.2 & 6.4 & 61.6 & 7.0 \\
		3   & 68.4 & 6.0 & 67.8 & 6.1 & 68.4 & 5.9 & 70.1 & 6.2 & 71.1 & 5.4 & 69.5 & 6.2 & 70.6 & 5.7 & 71.2 & 5.8 \\
		4   & 73.8 & 5.4 & 73.6 & 5.7 & 75.2 & 5.2 & 73.5 & 5.6 & 75.6 & 5.5 & 76.5 & 5.4 & 75.1 & 5.9 & 76.8 & 5.2 \\
		5   & 79.1 & 5.3 & 81.0 & 4.8 & 82.7 & 4.9 & 82.1 & 5.1 & 82.1 & 4.6 & 83.9 & 4.3 & 81.7 & 4.8 & 82.4 & 4.8 \\
		6   & 86.1 & 4.1 & 87.1 & 3.9 & 89.1 & 3.6 & 91.2 & 2.8 & 89.1 & 3.1 & 89.0 & 3.0 & 88.1 & 3.1 & 87.5 & 3.6 \\
		7   & 92.5 & 2.3 & 95.3 & 1.4 & 95.5 & 1.4 & 96.4 & 1.2 & 96.7 & 1.0 & 96.9 & 1.2 & 96.7 & 1.1 & 96.3 & 1.6 \\
		8   & 92.2 & 2.5 & 94.7 & 1.5 & 94.7 & 1.6 & 94.8 & 1.7 & 95.5 & 1.1 & 96.2 & 1.2 & 95.7 & 1.0 & 95.7 & 1.3 \\
		9   & 91.3 & 3.0 & 93.4 & 2.0 & 93.1 & 2.0 & 92.8 & 2.1 & 93.1 & 1.7 & 93.0 & 1.9 & 93.4 & 1.5 & 92.2 & 1.8 \\
		\midrule
		Sum & 90.6 & 3.2 & 93.3	& 2.2 & 93.6 & 2.1 & 94.0 & 2.2 & 94.8 & 1.6 & 95.2 & 1.6 & 95.3 & 1.2 & 94.2 & 1.8 \\
		\bottomrule
	\end{tabular}
\end{table*}

Analyzing performance across model complexities, it can be seen that features learned by less complex variants of EfficientNet (i.e.\ B0--B3, cf.\ Table~\ref{tab:features_mahalanobis_efficientnet}) perform worse in a transfer learning AD setting.
Further, it can be seen that the performance saturates eventually, and even degrades for EfficientNet-B7.
This could indicate that more complex EfficientNet variants start to overfit on ImageNet and no longer learn features that generalize well to new domains/use cases.
A similar effect is observed in our evaluations with Mahalanobis distance on ResNet architectures (cf. Appendix Table~\ref{tab:features_resnet_maha}).

\subsection{Why Pre-Trained Deep Features Work so Well}
Next, we investigate possible reasons for the oustanding performance of the MVG enacted in a transfer learning setting.
We hypothesize that features discriminative to the AD task do not necessarily vary strongly within the normal dataset, as implicitly presumed by semi-supervised approaches that employ normal data only. 
Therefore, we perform PCA on feature levels of a pre-trained EfficientNet-B4 and keep only principal components accounting for most of the variance before fitting the MVG to the dataset.
Vice versa, we retain principal components with the least amount of variance (i.e.\ those with smallest eigenvalues), which we denote as negated PCA (NPCA) in the following.

\begin{table*}
	\caption[AUROC ($\pm$ SEM) scores in percent for PCA and negated PCA compressed EfficientNet-B4 features with Mahalanobis distance]{AUROC ($\pm$SEM) scores in percent for negated PCA compressed EfficientNet-B4 features with Mahalanobis distance}
	\label{tab:features_npca}
	\centering
	\begin{tabular}{@{}crrrrrrrrrrrr@{}}
		\toprule
		\multirow{2}{*}{Level} & \multicolumn{2}{c}{No Compression} & \multicolumn{2}{c}{PCA 99\%} &  \multicolumn{2}{c}{PCA 95\%} & \multicolumn{2}{c}{NPCA 1\%} & \multicolumn{2}{c}{NPCA 0.1\%} & \multicolumn{2}{c@{}}{NPCA 0.01\%} \\
		\cmidrule(lr){2-3}
		\cmidrule(lr){4-5}
		\cmidrule(lr){6-7}
		\cmidrule(l){8-9}
		\cmidrule(l){10-11}
		\cmidrule(l){12-13}
		  & Mean & SEM & Mean & SEM & Mean & SEM & Mean & SEM & Mean & SEM & Mean & SEM \\
		\midrule
		1 & 60.3 & 6.1 & 50.8 & 6.1 & 45.8 & 5.4 & 64.1 & 6.4 & 67.8 & 6.6 & 69.9 & 6.2 \\
		2 & 62.0 & 6.4 & 53.3 & 5.9 & 48.6 & 5.7 & 67.6 & 6.3 & 68.0 & 5.9 & 67.3 & 5.7 \\
		3 & 71.1 & 5.4 & 65.5 & 6.5 & 59.7 & 6.2 & 71.6 & 4.9 & 68.1 & 4.2 & 65.7 & 3.7 \\
		4 & 75.6 & 5.5 & 69.5 & 6.1 & 63.2 & 6.4 & 76.1 & 5.1 & 73.1 & 4.6 & 69.4 & 4.0 \\
		5 & 82.1 & 4.6 & 76.2 & 5.3 & 66.6 & 6.5 & 82.5 & 4.0 & 78.7 & 3.6 & 72.3 & 3.7 \\
		6 & 89.1 & 3.1 & 83.3 & 4.8 & 77.3 & 5.7 & 90.2 & 2.5 & 88.2 & 2.4 & 83.6 & 2.9 \\
		7 & 96.7 & 1.0 & 93.4 & 2.1 & 87.1 & 4.0 & 96.1 & 1.0 & 94.5 & 1.3 & 89.6 & 2.5 \\
		8 & 95.5 & 1.1 & 91.9 & 2.1 & 88.6 & 3.1 & 94.8 & 1.2 & 93.8 & 1.4 & 90.6 & 2.3 \\
		9 & 93.1 & 1.7 & 91.3 & 2.1 & 88.6 & 2.9 & 93.3 & 1.6 & 91.2 & 2.1 & 86.3 & 3.0 \\
		\midrule
		Sum & 94.8 & 1.6 & 89.6 & 3.4 & 82.2 & 6.0 & 95.6 & 1.3 & 95.5 & 1.2 & 94.0 & 1.6 \\
		\bottomrule
	\end{tabular}
\end{table*}
When assessing effects of PCA-based dimensionality reduction two observations can be made.
First, keeping only principal components that retain high variance in the training datasets reduces AD performance.
In fact, removing principal components that account for a total of 1\% of variance leads to a reduction in AD performance across all levels (cf.\ PCA 99\% in Table~\ref{tab:features_npca}).
Conversely, when retaining only principal components with small eigenvalues, no considerable performance is lost and performance for the sum mode even increases (Table~\ref{tab:features_npca}). 
Therefore, NPCA offers an elegant way to reduce dimensionality of the pre-trained feature spaces.
For reference, NPCA 0.01\% reduces dimensionality of the feature space in level 7 from 272 to 15.6 features on average across all folds and categories.
This alleviates the curse of dimensionality and makes it easier to fit the MVG even with little training data. 

Furthermore, this finding supports the hypothesis that AD algorithms which learn features from scratch utilizing only normal data perform worse than AD approaches using pre-trained features. 
The reason is that feature combinations that retain little variance in normal data (i.e.\ do not occur in normal data and can thus not be learned effectively) are ultimately those that can be used to discriminate between normal and anomalous images.

\subsection{Choosing a Working Point Solely on FPR}

While our evaluation has focussed on AUROC, neglecting the issue of choosing a working point, the MVG assumption also offers a theoretical framework for selecting the working point by choosing an acceptable FPR out of the box (cf. (\ref{eq:threshold_fpr})).
Note that a target FPR cannot be easily set for the sum mode where the Mahalanobis distances of feature-level MVGs are added.
Therefore, we restrict our evaluations to level 7 features of two different EfficientNet variants, choosing EfficientNet-B0 for its low model complexity and EfficientNet-B4 for its high AD performance at medium complexity.
We assess effects of performing no compression, 99\% PCA and 0.01\% NPCA compression.
Here, we compare target FPR based on (\ref{eq:threshold_fpr}) to the FPR achieved on the test set, also reporting the TPR yielded by that working point and overall AUROC.
We also assess the potentially beneficial effect of augmentations in order to to artificially enlarge small datasets for more robust covariance estimation.
Augmentations are selected per MVTec category to avoid accidental transformation of normal to anomalous data (details can be found in the Appendix Fig.~\ref{fig:augmentation_pipeline}).
We artifically increase each dataset's size by aggregating over 100 epochs.

\begin{table}
	\caption{Expected FPR and achieved FPR / TPR in percent per multiple $n\cdot \sigma$ for EfficientNet level 7 features under different compression modes.
	AUROC values are also reported.}
	\label{tab:thresholds_sigma_level7}
	\centering
	\renewcommand{\tabcolsep}{5.1pt} 
	\begin{tabular}{@{}l
		S[round-integer-to-decimal,
		  table-format=2.0e-1,
		  tight-spacing=true]
		S[round-integer-to-decimal,
		  table-format=2.1,
		  tight-spacing=true]
		S[round-integer-to-decimal,
		  table-format=2.1,
		  tight-spacing=true]
		S[round-integer-to-decimal,
		  table-format=2.1,
		  tight-spacing=true]
		S[round-integer-to-decimal,
		  table-format=2.1,
		  tight-spacing=true]
		S[round-integer-to-decimal,
		  table-format=2.1,
		  tight-spacing=true]
		S[round-integer-to-decimal,
		  table-format=2.1,
		  tight-spacing=true]
		S[round-integer-to-decimal,
		  table-format=2.1,
		  tight-spacing=true]
		S[round-integer-to-decimal,
		  table-format=2.1,
		  tight-spacing=true]
		@{}}
		\toprule
		{\multirow{3}{*}{$n$}}	&	{\multirow{3}{*}{\shortstack[c]{Target\\FPR}}}	& \multicolumn{6}{c}{EN-B0} & \multicolumn{2}{c@{}}{EN-B4} \\
		\cmidrule(l{0.6375em}r{0.6375em}){3-8}
		\cmidrule(l{0.6375em}){9-10}
		& 	 & \multicolumn{2}{c}{PCA 99\%} & \multicolumn{2}{c}{All Features} & \multicolumn{2}{c}{NPCA 1\%} & \multicolumn{2}{c@{}}{All Features} \\
		\cmidrule(l{0.6375em}r{0.6375em}){3-4}
		\cmidrule(l{0.6375em}r{0.6375em}){5-6}
		\cmidrule(l{0.6375em}r{0.6375em}){7-8}
		\cmidrule(l{0.6375em}){9-10}
				& 	 				 & {FPR} & {TPR} & {FPR} & {TPR} & {FPR} & {TPR} & {FPR}  & {TPR} \\
		\midrule
		1 		& 31.7 				 & 17.9	& 77.8	 & 30.4	 & 89.1	 &  47.4 & 94.0  & 66.1   & 98.4 \\
		2 		& 4.6 				 & 10.1	& 71.2	 & 19.3	 & 84.2	 &  29.8 & 88.9  & 56.0   & 97.3 \\
		3 		& 0.3 				 & 5.8	& 66.1	 & 13.5	 & 80.2	 &  18.9 & 83.2  & 44.7   & 95.9 \\
		4 		& 6E-3				 & 1.9	& 61.3	 & 9.3	 & 76.3	 &  11.3 & 78.5  & 35.7   & 94.0 \\
		5 		& 6E-5				 & 0.7	& 57.0	 & 6.5	 & 72.7	 &  6.8  & 74.4  & 28.2   & 91.6 \\
		\midrule
		\multicolumn{2}{@{}l}{AUROC} & \multicolumn{2}{c}{90.3}  	& \multicolumn{2}{c}{93.2} 	& \multicolumn{2}{c}{93.7}	& \multicolumn{2}{c@{}}{97.1}		\\
		\bottomrule
	\end{tabular}
\end{table}

Performing the experiments, we observsed that augmentations were essential to enable setting the working point that completely failed otherwise (e.g.\ FPR of 99.8\% and TPR of 99.9\% were achieved 3$\sigma$ for EfficientNet-B0 at no compression).
Looking at Table~\ref{tab:thresholds_sigma_level7}, it can be observed that PCA decreases FPRs yielded on the test set, whereas NPCA increases the FPRs.
Therefore, PCA and NPCA behave inverse to each other, and PCA compression may prove useful in providing robust estimates of achieved FPRs at the cost of reduced AD performance.
Furthermore, even with artificially enlarged datasets, a sensible setting of the FPR based on training data is possible only for the smallest model, EfficientNet-B0.
This indicates the curse of dimensionality, as complex models require increasingly more data to avoid overfitting on noise present in the training data (bias-variance tradeoff).
Thus, experiments should be reevaluated on larger AD datasets to confirm our findings.
Still, setting the working point by means of a FPR can be realized via this theoretical framework. 
This is novel, as purely empirical approaches dominate the current literature (i.e.\ setting working point based on a hold-out validation set before applying to the test set).


\subsection{Comparison with State of the Art on MVTec AD}

Finanlly, we compare the performance of our proposed AD approach with state-of-the-art AD algorithms on the MVTec AD dataset.
Our evaluation comprises a semi-supervised reconstruction approach using a convolutional AE, a fully-supervised AD classifier as well as an oc-SVM fit to the pre-trained feature representations of EfficientNet.
We further compare to the current state-of-the-art performance reported in literature on MVTec, taking the correspoding values directly from the linked sources.

While the fully-supervised classifier can not be deployed in practice to AD problems, it serves as an upper bound of what can be achieved by AD algorithms.
Here, we fine-tune a pre-trained EfficientNet-B4, EfficientNet-B2 as well as ResNet-18 and ResNet-34 variants per category.
For data splits, we still perform a 5-fold evaluation, but no longer adhere to the original MVTec splits, as there are no anomalies present in the train datasets.
Instead, we pool both train and test set and stratify splits to maintain identical anomaly prevalence in all folds.
We compute AUROC on a val set split from the train set to select the best model state.
The best model is then applied to the unused test set.
For training, we select and apply the same augmentations per-category as used to artficially enlarge dataset size (Appendix Fig.~\ref{fig:augmentation_pipeline}). 
We use a batch-size of 64 for ResNet, 16 for EfficientNet and train using the Adam \cite{Kingma2015} optimizer with an initial learning rate of 0.0001 employing the binary cross-entropy loss function. 

For the AE, we choose the ResNet-18 for the encoder and an inverted ResNet-18 for the decoder part (i.e.\ every operation of the encoder should be inverted by the decoder).
For the upsampling operations we employ pixel shuffle operations as introduced by Shi et al.\ \cite{Shi2016a} to reduce checkerboard artifacts which would be present otherwise.
The latent dimension of the bottleneck is set to 32 and yields proper reconstruction of the normal class in all categories.
We also generally employ the same augmentations as before, but disable noise augmentations (cf.\ Appendix Fig.~\ref{fig:augmentation_pipeline}).
Batch-size, learning rate and optimizer are the same as for the supervised classifier, and based on preliminary experiments the $L_2$-distance is chosen for the reconstruction error.
As stated before, an aggregation of the residual image to image-level information is necessary for reconstruction-based approaches.
While the threshold employed for ROC calculation is set on the pixel level, we perform connected component analysis and label a test image as defective only if it contains a connected component at least as big as the smallest anomaly present in the test dataset.
Note that by extracting the minimal anomaly size from the test dataset, knowledge about the process is introduced to the AE approach, increasing complexity of the procedure.

To enhance comparability, we also apply augmentation to covariance estimation and compute features across 100 epochs of normal training data per split.
To demonstrate the general applicability of our approach, we evaluate the proposed approach in sum mode over all feature levels of EfficientNet-B4, as we can achieve comparable performance and further reduce complexity by omitting feature level selection.

For the oc-SVM, we fit a RBF-kernel model to every feature level using normal data only and aggregate the predicted anomaly score over all levels to yield a sum score similar to the proposed pipeline. 

\begin{table} 
	\caption{Comparison to the state of the art. Values for state-of-the-art methods are directly taken from the corresponding sources.
	We report AUROC ($\pm$SEM) scores in percent.
	The AE approaches map-mean and CCA stand for score map mean and connected component analysis, respectively.
	Mahalanobis and oc-SVM approaches are summed over all feature levels.
	The highest AUROC amongst non-fully supervised methods is boldfaced.}
	\label{tab:final_results_literature}
	\centering
	\begin{tabular}{@{}p{1em}@{}llrr@{}}
		\toprule
		\multicolumn{2}{@{}c}{Approach} & Architecture & Mean & SEM \\
		\midrule
		\multicolumn{2}{@{}l}{GeoTrans \cite{Golan2018} (source: \cite{Huang2019})} & Wide-ResNet & 67.2 & 4.7 \\
		\multicolumn{2}{@{}l}{GANomaly \cite{Akcay2018} (source: \cite{Huang2019})} & DCGAN & 76.1 & 1.6 \\
		\multicolumn{2}{@{}l}{ITAE \cite{Huang2019}} & Custom & 83.9 & 2.8 \\
		\multicolumn{2}{@{}l}{SPADE \cite{Cohen2020}} & Wide-ResNet50-2 & 85.5 & {---}\\
		\midrule
		\multicolumn{5}{@{}l}{MSE AE } \\
		& Map-Mean & ResNet-18 & 78.8 & 4.1 \\
		& CCA & ResNet-18 & 81.8 & 3.4 \\
		\multicolumn{5}{@{}l}{Pre-Trained Classifier} \\
		& Fully-Supervised Fine-tune & ResNet-18 & 93.3 & 1.4 \\
		& Fully-Supervised Fine-tune & ResNet-34 & 93.4 & 1.3 \\
		& Fully-Supervised Fine-tune & EfficientNet-B0 & 94.1 & 1.4	\\
		& Fully-Supervised Fine-tune & EfficientNet-B4 & 96.3 & 1.0 \\
		\multicolumn{5}{@{}l}{Oc-SVM} \\
		& All Features & EfficientNet-B0 & 73.0	& 6.1 \\
		& All Features & EfficientNet-B4 & 78.1 & 4.7 \\
		\multicolumn{5}{@{}l}{Mahalanobis (ours)} \\
		& All Features & EfficientNet-B4 & 95.2 & 1.5 \\
		& NPCA 1\% & EfficientNet-B4 & \textbf{95.8} & 1.2 \\
		\bottomrule
	\end{tabular}
\end{table}

Assessing performance results, it becomes apparent that MVG estimation on pre-trained deep features vastly outperforms the state of the art on MVTec, achieving 10\% higher average AUROC than the next best model SPADE (cf.\ Table~\ref{tab:final_results_literature}).
Notably, SPADE as proposed by Cohen et al. \cite{Cohen2020} also leverages pre-trained deep feature spaces in combination with deep $k$-NN and $L_2$ loss. 
Also, NPCA slightly improves robustness of the method, leading to an increased average AUROC score and a decreased SEM. 
Furthermore, performance is mostly comparable to fully-supervised fine-tuning of pre-trained classifiers but at times slightly worse (cf. per-category results reported in Appendix Table~\ref{tab:final_results_categories}).
This is especially the case for the texture categories, whereas for some object categories (e.g.\ \textit{pill}, \textit{screw}) better performance is achieved by fine-tuning.
We stress again that fully-supervised AD can not be realistically applied to most AD problems, and only serves as an upper limit of achievable AUROC performance.

\section{Discussion}
We have demonstrated the benefits of using ImageNet pre-training for general-purpose AD in images.
In particular, we showed that the MVG assumption of high correlation in pre-trained deep features is crucial to attain state-of-the-art AD performance (cf.\ Table~\ref{tab:features_comparison_efficientnet}).
Therefore, the generative assumption proposed by Lee et al.\ \cite{Lee2018} holds also in a transfer learning setting for AD in images semantically different to the training dataset and can be leveraged to improve AD.
Our PCA analysis revealed that discriminative components of the transferred deep feature representations contain little overall variance in normal data and should thus be difficult to learn.
This finding is in agreement with literature on shallow AD \cite{Tax2003} and explains the poor performance of OC-SVM when applied to deep feature representations \cite{MefrazKhan2014}.
We therefore agree with Bergman et al.\ \cite{Bergman2020a} that pre-trained feature spaces should be adopted for AD in a transfer learning approach instead of learning features from scratch using normal data only.
We further expand upon this and conclude that the MVG assumption is crucial to fully recapitulate the characteristics of pre-trained deep feature representations and to realize their potential.

While good performance has been achieved on the MVTec AD dataset, we expect that with increasing semantic distance to natural images (e.g.\ images of the medical domain), out-of-the-box AD performance will decrease.
In such scenarios, pre-trained models could be used as starting points for fine-tuning on target domains, e.g.\ by employing the deep SVDD proposed by Ruff et al.\ \cite{Ruff2018}.
The MVG assumption can be easily integrated here, as the hypersphere could be initialized by transforming the features with the inverse Cholesky decomposition of the estimated covariance. 

Apart from the unimodal setting, AD may also occur in a multi-modal context \cite{Ahmed2019}.
We therefore will extend the presented AD approach to multi-modal distributions on a complex in-house fabric dataset (e.g.\ by fitting Gaussian Mixture Models).

The MVG further offers a framework to set the working point by determining an acceptable FPR.
However, while augmentations and low model complexities were shown to alleviate the mismatch between desired and achieved FPR, an evaluation on even larger AD datasets is required.
Further, modifications to the model architecture may prove beneficial to enforce a normal distribution in deep feature representations, as we have seen from the difference in performance between ResNet and EfficientNet features.
Here, Self-Normalizing Neural Networks (SNNs) provide a basis for learning Gaussian-distributed features \cite{Klambauer2017}.
Also, class-wise distributions may be explicitly constrained to follow Gaussians in deep feature spaces during ImageNet pretraining.
It should be noted that while the FPR may be selected, Probably Approximately Correct (PAC) style TPR guarantees, as required for security-critical use-cases, cannot be given by this approach and are a different field of research.
Here, Liu et al. \cite{Liu2018} achieved PAC-style guarantees for the TPR, requiring well-defined anomaly distributions to do so.

\section{Conclusion}

We have achieved state-of-the-art performance on the public MVTec AD dataset using deep feature representations extracted from classifiers pre-trained on ImageNet.
Our approach is simple yet effective and consists of fitting a MVG to normal data in the pre-trained deep feature representations, using Mahalanobis distance as anomaly score.
We have further investigated the reason behind the effectiveness of our approach.
Using PCA, we reveal that principal components containing only little variance in normal data are ultimately those necessary for discriminating between normal and anomalous images. 
We argue that these features are difficult to learn from scratch using normal data only, and propose to instead use feature representations generated by large-scale discriminative training in a transfer learning setting assuming MVG distributions.
Future research in image AD should focus on (I) increasing the generalizability of pre-trained features, (II) fine-tuning of transferred representations using the small available datasets as well as (III) extending the approach to AD tasks with multi-modal normal data distributions. 






\bibliographystyle{./template/IEEEtran}
\bibliography{IEEEabrv,./Literatur/lit}

\begin{thebibliography}{10}
\providecommand{\url}[1]{#1}
\csname url@samestyle\endcsname
\providecommand{\newblock}{\relax}
\providecommand{\bibinfo}[2]{#2}
\providecommand{\BIBentrySTDinterwordspacing}{\spaceskip=0pt\relax}
\providecommand{\BIBentryALTinterwordstretchfactor}{4}
\providecommand{\BIBentryALTinterwordspacing}{\spaceskip=\fontdimen2\font plus
\BIBentryALTinterwordstretchfactor\fontdimen3\font minus
  \fontdimen4\font\relax}
\providecommand{\BIBforeignlanguage}[2]{{%
\expandafter\ifx\csname l@#1\endcsname\relax
\typeout{** WARNING: IEEEtran.bst: No hyphenation pattern has been}%
\typeout{** loaded for the language `#1'. Using the pattern for}%
\typeout{** the default language instead.}%
\else
\language=\csname l@#1\endcsname
\fi
#2}}
\providecommand{\BIBdecl}{\relax}
\BIBdecl

\bibitem{Chandola2009}
V.~Chandola, A.~Banerjee, and V.~Kumar, ``Anomaly detection: A survey,''
  \emph{ACM computing surveys (CSUR)}, vol.~41, no.~3, pp. 1--58, 2009.

\bibitem{Pimentel2014}
M.~A. Pimentel, D.~A. Clifton, L.~Clifton, and L.~Tarassenko, ``A review of
  novelty detection,'' \emph{Signal Processing}, vol.~99, pp. 215--249, 2014.

\bibitem{Bergmann2019}
P.~Bergmann, M.~Fauser, D.~Sattlegger, and C.~Steger, ``{MVTec AD} -- a
  comprehensive real-world dataset for unsupervised anomaly detection,'' in
  \emph{The IEEE Conference on Computer Vision and Pattern Recognition (CVPR)},
  6 2019.

\bibitem{Schlegl2019}
T.~Schlegl, P.~Seeb{\"o}ck, S.~M. Waldstein, G.~Langs, and U.~Schmidt-Erfurth,
  ``{f-AnoGAN}: Fast unsupervised anomaly detection with generative adversarial
  networks,'' \emph{Medical image analysis}, vol.~54, pp. 30--44, 2019.

\bibitem{Ruff2018}
L.~Ruff, R.~Vandermeulen, N.~Goernitz, L.~Deecke, S.~A. Siddiqui, A.~Binder,
  E.~M{\"u}ller, and M.~Kloft, ``Deep one-class classification,'' in
  \emph{Proceedings of the 35th International Conference on Machine Learning},
  ser. Proceedings of Machine Learning Research, J.~Dy and A.~Krause, Eds.,
  vol.~80.\hskip 1em plus 0.5em minus 0.4em\relax Stockholmsmässan, Stockholm
  Sweden: PMLR, 10--15 Jul 2018, pp. 4393--4402.

\bibitem{Deng2009}
J.~Deng, W.~Dong, R.~Socher, L.~Li, K.~Li, and L.~Fei-Fei, ``{ImageNet}: A
  large-scale hierarchical image database,'' in \emph{2009 IEEE Conference on
  Computer Vision and Pattern Recognition}, June 2009, pp. 248--255.

\bibitem{Napoletano2018}
P.~Napoletano, F.~Piccoli, and R.~Schettini, ``Anomaly detection in nanofibrous
  materials by {CNN}-based self-similarity,'' \emph{Sensors}, vol.~18, no.~1,
  p. 209, 2018.

\bibitem{Andrews2016}
J.~Andrews, T.~Tanay, E.~J. Morton, and L.~D. Griffin, ``Transfer
  representation-learning for anomaly detection,'' JMLR.\hskip 1em plus 0.5em
  minus 0.4em\relax Multidisciplinary Digital Publishing Institute, 2016.

\bibitem{Cohen2020}
N.~Cohen and Y.~Hoshen, ``Sub-image anomaly detection with deep pyramid
  correspondences,'' \emph{arXiv preprint arXiv:2005.02357}, 2020.

\bibitem{Gong2019}
D.~Gong, L.~Liu, V.~Le, B.~Saha, M.~R. Mansour, S.~Venkatesh, and A.~v.~d.
  Hengel, ``Memorizing normality to detect anomaly: Memory-augmented deep
  autoencoder for unsupervised anomaly detection,'' in \emph{Proceedings of the
  IEEE International Conference on Computer Vision}, 2019, pp. 1705--1714.

\bibitem{Mahalanobis1936}
P.~C. Mahalanobis, ``On the generalized distance in statistics.''\hskip 1em
  plus 0.5em minus 0.4em\relax National Institute of Science of India, 1936.

\bibitem{Bergmann2018}
P.~Bergmann, S.~L{\"o}we, M.~Fauser, D.~Sattlegger, and C.~Steger, ``Improving
  unsupervised defect segmentation by applying structural similarity to
  autoencoders,'' \emph{arXiv preprint arXiv:1807.02011}, 2018.

\bibitem{Haselmann2018}
M.~Haselmann, D.~P. Gruber, and P.~Tabatabai, ``Anomaly detection using deep
  learning based image completion,'' in \emph{2018 17th IEEE International
  Conference on Machine Learning and Applications (ICMLA)}.\hskip 1em plus
  0.5em minus 0.4em\relax IEEE, 2018, pp. 1237--1242.

\bibitem{Daniel2019}
T.~Daniel, T.~Kurutach, and A.~Tamar, ``Deep variational semi-supervised
  novelty detection,'' \emph{arXiv preprint arXiv:1911.04971}, 2019.

\bibitem{Pidhorskyi2018}
S.~Pidhorskyi, R.~Almohsen, and G.~Doretto, ``Generative probabilistic novelty
  detection with adversarial autoencoders,'' in \emph{Advances in neural
  information processing systems}, 2018, pp. 6822--6833.

\bibitem{Schoelkopf2001}
B.~Sch{\"o}lkopf, J.~C. Platt, J.~Shawe-Taylor, A.~J. Smola, and R.~C.
  Williamson, ``Estimating the support of a high-dimensional distribution,''
  \emph{Neural computation}, vol.~13, no.~7, pp. 1443--1471, 2001.

\bibitem{Sarafijanovic-Djukic2019}
N.~Sarafijanovic-Djukic and J.~Davis, ``Fast distance-based anomaly detection
  in images using an inception-like autoencoder,'' in \emph{Discovery Science},
  P.~Kralj~Novak, T.~{\v{S}}muc, and S.~D{\v{z}}eroski, Eds.\hskip 1em plus
  0.5em minus 0.4em\relax Cham: Springer International Publishing, 2019, pp.
  493--508.

\bibitem{Ruff2020}
L.~Ruff, R.~A. Vandermeulen, N.~Görnitz, A.~Binder, E.~Müller, K.-R. Müller,
  and M.~Kloft, ``Deep semi-supervised anomaly detection,'' in
  \emph{International Conference on Learning Representations}, 2020.

\bibitem{Abati2019}
D.~Abati, A.~Porrello, S.~Calderara, and R.~Cucchiara, ``Latent space
  autoregression for novelty detection,'' in \emph{The IEEE Conference on
  Computer Vision and Pattern Recognition (CVPR)}, June 2019.

\bibitem{Zong2018}
B.~Zong, Q.~Song, M.~R. Min, W.~Cheng, C.~Lumezanu, D.~Cho, and H.~Chen, ``Deep
  autoencoding gaussian mixture model for unsupervised anomaly detection,'' in
  \emph{International Conference on Learning Representations}, 2018.

\bibitem{Vasilev2018}
A.~Vasilev, V.~Golkov, M.~Meissner, I.~Lipp, E.~Sgarlata, V.~Tomassini, D.~K.
  Jones, and D.~Cremers, ``q-space novelty detection with variational
  autoencoders,'' \emph{arXiv preprint arXiv:1806.02997}, 2018.

\bibitem{Sabokrou2018}
M.~Sabokrou, M.~Fayyaz, M.~Fathy, Z.~Moayed, and R.~Klette, ``Deep-anomaly:
  Fully convolutional neural network for fast anomaly detection in crowded
  scenes,'' \emph{Computer Vision and Image Understanding}, vol. 172, pp.
  88--97, 2018.

\bibitem{Bergmann2020}
P.~Bergmann, M.~Fauser, D.~Sattlegger, and C.~Steger, ``Uninformed students:
  Student-teacher anomaly detection with discriminative latent embeddings,'' in
  \emph{Proceedings of the IEEE/CVF Conference on Computer Vision and Pattern
  Recognition}, 2020, pp. 4183--4192.

\bibitem{Christiansen2016}
P.~Christiansen, L.~N. Nielsen, K.~A. Steen, R.~N. J{\o}rgensen, and
  H.~Karstoft, ``Deepanomaly: Combining background subtraction and deep
  learning for detecting obstacles and anomalies in an agricultural field,''
  \emph{Sensors}, vol.~16, no.~11, p. 1904, 2016.

\bibitem{Krizhevsky2012}
A.~Krizhevsky, I.~Sutskever, and G.~E. Hinton, ``{ImageNet} classification with
  deep convolutional neural networks,'' in \emph{Advances in neural information
  processing systems}, 2012, pp. 1097--1105.

\bibitem{Simonyan2015}
K.~Simonyan and A.~Zisserman, ``Very deep convolutional networks for
  large-scale image recognition,'' in \emph{3rd International Conference on
  Learning Representations, {ICLR} 2015, San Diego, CA, USA, May 7-9, 2015,
  Conference Track Proceedings}, Y.~Bengio and Y.~LeCun, Eds., 2015.

\bibitem{Bergman2020a}
L.~Bergman, N.~Cohen, and Y.~Hoshen, ``Deep nearest neighbor anomaly
  detection,'' \emph{arXiv preprint arXiv:2002.10445}, 2020.

\bibitem{He2016}
K.~He, X.~Zhang, S.~Ren, and J.~Sun, ``Deep residual learning for image
  recognition,'' in \emph{Proceedings of the IEEE conference on computer vision
  and pattern recognition}, 2016, pp. 770--778.

\bibitem{Nazare2018}
T.~S. Nazare, R.~F. de~Mello, and M.~A. Ponti, ``Are pre-trained {CNN}s good
  feature extractors for anomaly detection in surveillance videos?''
  \emph{arXiv preprint arXiv:1811.08495}, 2018.

\bibitem{Lee2018}
K.~Lee, K.~Lee, H.~Lee, and J.~Shin, ``A simple unified framework for detecting
  out-of-distribution samples and adversarial attacks,'' in \emph{Advances in
  Neural Information Processing Systems}, 2018, pp. 7167--7177.

\bibitem{Ledoit2004}
O.~Ledoit, M.~Wolf \emph{et~al.}, ``A well-conditioned estimator for
  large-dimensional covariance matrices,'' \emph{Journal of Multivariate
  Analysis}, vol.~88, no.~2, pp. 365--411, 2004.

\bibitem{Tan2019}
M.~Tan and Q.~V. Le, ``{EfficientNet}: Rethinking model scaling for
  convolutional neural networks,'' in \emph{Proceedings of the 36th
  International Conference on Machine Learning, {ICML} 2019, 9-15 June 2019,
  Long Beach, California, {USA}}, ser. Proceedings of Machine Learning
  Research, K.~Chaudhuri and R.~Salakhutdinov, Eds., vol.~97.\hskip 1em plus
  0.5em minus 0.4em\relax {PMLR}, 2019, pp. 6105--6114.

\bibitem{Ferri2011}
C.~Ferri, J.~Hern{\'a}ndez-Orallo, and P.~A. Flach, ``A coherent interpretation
  of {AUC} as a measure of aggregated classification performance,'' in
  \emph{Proceedings of the 28th International Conference on Machine Learning
  (ICML-11)}, 2011, pp. 657--664.

\bibitem{Ramachandran2017}
P.~Ramachandran, B.~Zoph, and Q.~V. Le, ``Searching for activation functions,''
  \emph{arXiv preprint arXiv:1710.05941}, 2017.

\bibitem{Hsu2020}
Y.-C. Hsu, Y.~Shen, H.~Jin, and Z.~Kira, ``Generalized odin: Detecting
  out-of-distribution image without learning from out-of-distribution data,''
  in \emph{Proceedings of the IEEE/CVF Conference on Computer Vision and
  Pattern Recognition (CVPR)}, June 2020.

\bibitem{Kingma2015}
D.~P. Kingma and J.~Ba, ``Adam: {A} method for stochastic optimization,'' in
  \emph{3rd International Conference on Learning Representations, {ICLR} 2015,
  San Diego, CA, USA, May 7-9, 2015, Conference Track Proceedings}, Y.~Bengio
  and Y.~LeCun, Eds., 2015.

\bibitem{Shi2016a}
W.~Shi, J.~Caballero, L.~Theis, F.~Huszar, A.~Aitken, C.~Ledig, and Z.~Wang,
  ``Is the deconvolution layer the same as a convolutional layer?'' \emph{arXiv
  preprint arXiv:1609.07009}, 2016.

\bibitem{Golan2018}
I.~Golan and R.~El-Yaniv, ``Deep anomaly detection using geometric
  transformations,'' in \emph{Advances in Neural Information Processing
  Systems}, 2018, pp. 9758--9769.

\bibitem{Huang2019}
C.~Huang, J.~Cao, F.~Ye, M.~Li, Y.~Zhang, and C.~Lu, ``Inverse-transform
  autoencoder for anomaly detection,'' \emph{arXiv preprint arXiv:1911.10676},
  2019.

\bibitem{Akcay2018}
S.~Ak{\c{c}}ay, A.~Atapour-Abarghouei, and T.~P. Breckon, ``{GANomaly}:
  Semi-supervised anomaly detection via adversarial training,'' in \emph{Asian
  Conference on Computer Vision}.\hskip 1em plus 0.5em minus 0.4em\relax
  Springer, 2018, pp. 622--637.

\bibitem{Tax2003}
D.~M. Tax and K.-R. M{\"u}ller, ``Feature extraction for one-class
  classification,'' in \emph{Artificial Neural Networks and Neural Information
  Processing—ICANN/ICONIP 2003}.\hskip 1em plus 0.5em minus 0.4em\relax
  Springer, 2003, pp. 342--349.

\bibitem{MefrazKhan2014}
N.~{Mefraz Khan}, R.~Ksantini, I.~{Shafiq Ahmad}, and L.~Guan,
  ``Covariance-guided one-class support vector machine,'' \emph{Pattern
  Recognition}, vol.~47, no.~6, pp. 2165 -- 2177, 2014.

\bibitem{Ahmed2019}
F.~Ahmed and A.~Courville, ``Detecting semantic anomalies,'' \emph{arXiv
  preprint arXiv:1908.04388}, 2019.

\bibitem{Klambauer2017}
G.~Klambauer, T.~Unterthiner, A.~Mayr, and S.~Hochreiter, ``Self-normalizing
  neural networks,'' in \emph{Advances in Neural Information Processing Systems
  30}, I.~Guyon, U.~V. Luxburg, S.~Bengio, H.~Wallach, R.~Fergus,
  S.~Vishwanathan, and R.~Garnett, Eds.\hskip 1em plus 0.5em minus 0.4em\relax
  Curran Associates, Inc., 2017, pp. 971--980.

\bibitem{Liu2018}
S.~Liu, R.~Garrepalli, T.~Dietterich, A.~Fern, and D.~Hendrycks, ``Open
  category detection with {PAC} guarantees,'' in \emph{International Conference
  on Machine Learning}, 2018, pp. 3169--3178.

\end{thebibliography}
%
%
%

\cleardoublepage
\appendix

\begin{table}[htbp]
	\caption[EfficientNet-B0 baseline network]{EfficientNet-B0 baseline network (source: \cite{Tan2019})}
	\label{tab:efficient-b0_baseline}
	\centering
	\begin{tabular}{@{}clcrr@{}}
		\toprule
		Stage & Operator & Resolution & \#Channels & \#Layers \\
		$i$ & $F_i$ & $H_i \times W_i$ & $C_i$ & $L_i$ \\
		\midrule
		1 & Conv3x3 & $224 \times 224$ & 32 & 1 \\
		2 & MBConv1, k3x3 & $112 \times 112$ & 16 & 1 \\
		3 & MBConv6, k3x3 & $112 \times 112$ & 24 & 2 \\
		4 & MBConv6, k5x5 & $56 \times 56$ & 40 & 2 \\
		5 & MBConv6, k3x3 & $28 \times 28$ & 80 & 3 \\
		6 & MBConv6, k5x5 & $14 \times 14$ & 112 & 3 \\
		7 & MBConv6, k5x5 & $14 \times 14$ & 192 & 4 \\
		8 & MBConv6, k3x3 & $7 \times 7$ & 320 & 1 \\
    9 & Conv1x1 \& pool \& FC  & $7 \times 7$ & 1280 & 1 \\
    \bottomrule
	\end{tabular}
\end{table}

\begin{table}[htbp]
	\caption{Feature level AUROC ($\pm$ SEM) scores in percent for ResNet architectures with Mahalanobis distance}
	\label{tab:features_resnet_maha}
	\centering
	\begin{tabular}{@{}crrrrrr@{}}
		\toprule
		\multirow{2}{*}{Level} & \multicolumn{2}{c}{ResNet-18} & \multicolumn{2}{c}{ResNet-34} & \multicolumn{2}{c@{}}{ResNet-50} \\
		\cmidrule(lr){2-3}
		\cmidrule(lr){4-5}
		\cmidrule(l){6-7}
		& Mean & SEM & Mean & SEM & Mean & SEM \\
		\midrule
		1     & 66.6 & 6.4 & 67.5 & 6.8 & 68.0 & 5.9 \\
		2    & 71.6 & 6.4 & 71.9 & 6.1 & 73.7 & 6.4 \\
		3    & 78.6 & 5.0 & 79.4 & 4.8 & 81.3 & 4.9 \\
		4    & 86.7 & 4.1 & 90.4 & 3.6 & 89.0 & 4.3 \\
		5    & 88.3 & 2.7 & 89.0 & 3.0 & 86.9 & 3.8 \\
		\midrule
		Sum   & 86.4 & 4.0 & 88.2 & 4.0 & 87.3 & 4.5 \\
		\bottomrule
	\end{tabular}
\end{table}

\tikzset{%
	>={Latex[width=2mm,length=2mm]},
	base/.style = {rectangle, rounded corners, draw=black,
		minimum width=3cm, minimum height=1cm,
		text centered, font=\sffamily},
	data/.style = {base, fill=green},
	app/.style = {base, fill=orange},
	tool/.style = {base, fill=purple!70},
	process/.style = {base, minimum width=2.5cm, fill=red!50, font=\ttfamily},
}

\newcommand{\cbox}[1]{\raisebox{\depth}{\fcolorbox{black}{#1}{\null}}}
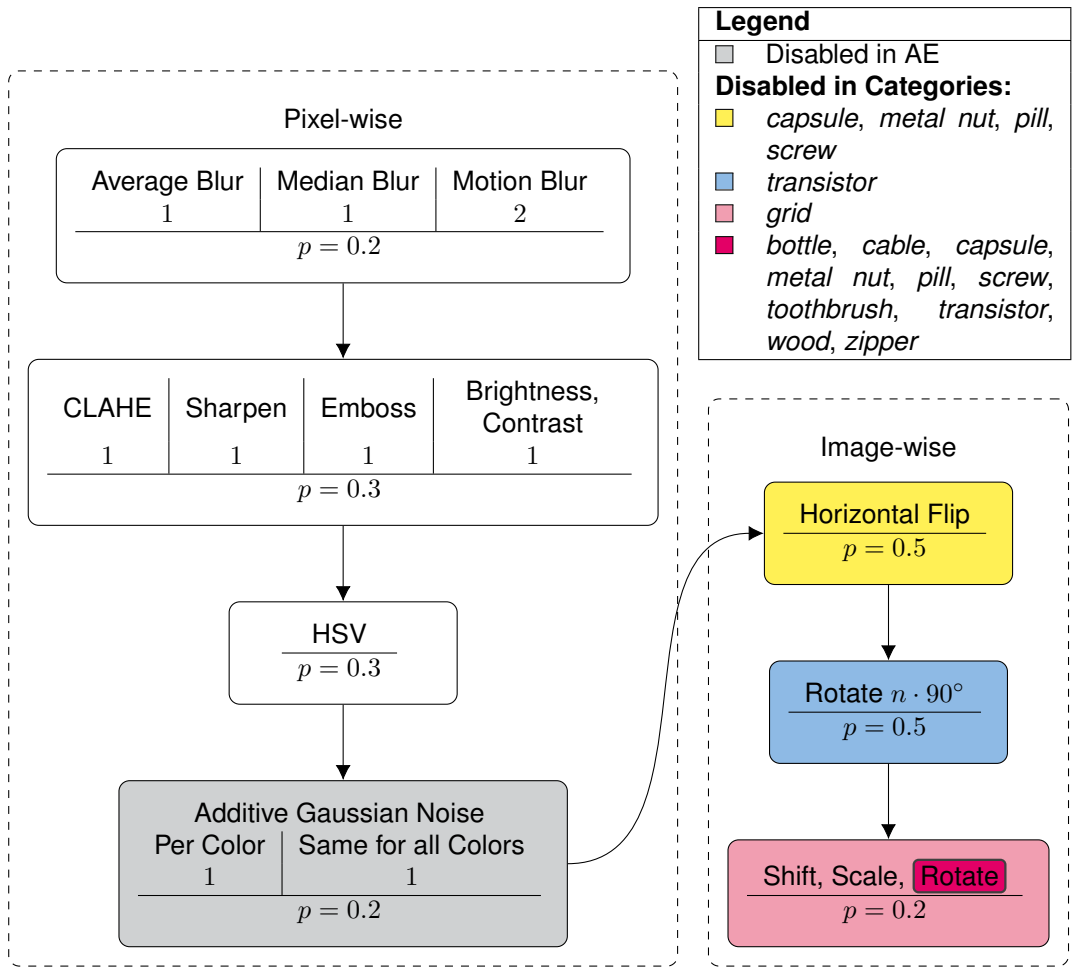
\begin{figure*}[htbp]
	\centering
	\begin{tikzpicture}[every node/.style={fill=white, font=\sffamily, inner sep=0.25cm}, align=center]

	\node (pixelwise) [] {Pixel-wise};
	\node (blur)   [base, below = of pixelwise, yshift=1cm] {
		\begin{tabular}{c|c|c}
		Average Blur & Median Blur & Motion Blur \\
		\num{1} & \num{1} & \num{2}\\\hline
		\multicolumn{3}{c}{$p=0.2$}
		\end{tabular}
	};
	\node (brightness)   [base, below = of blur] {
		\begin{tabular}{c|c|c|c}
		CLAHE & Sharpen & Emboss & \begin{tabular}{c}Brightness,\\ Contrast\end{tabular}\\
		\num{1} & \num{1} & \num{1} & \num{1}\\\hline
		\multicolumn{4}{c}{$p=0.3$}
		\end{tabular}
	};
	\node (hsv) [base, below = of brightness] {
		\begin{tabular}{c}
		HSV \\\hline
		$p=0.3$
		\end{tabular}
	};
	\node (noise)   [base, below = of hsv, fill=black-25] {
		\begin{tabular}{c|c}
		\multicolumn{2}{c}{Additive Gaussian Noise} \\
		Per Color & Same for all Colors \\
		\num{1} & \num{1}\\\hline
		\multicolumn{2}{c}{$p=0.2$}
		\end{tabular}
	};

	\node (shift) [base, right = of noise.south east, anchor=south west, xshift=1.1cm, fill=magenta-50] {
		\begin{tabular}{c}
		Shift, Scale, 
		\tcbox[colback=magenta,
			box align=base,
			size=small,
			left=0pt,
			right=0pt,
			boxsep=2pt,nobeforeafter]{Rotate} \\\hline
		$p=0.2$
		\end{tabular}
	};
	\node (rotate) [base, above = of shift, fill=rwth-50] {
		\begin{tabular}{c}
		Rotate $n\cdot 90^\circ$ \\\hline
		$p=0.5$
		\end{tabular}
	};
	\node (flip) [base, above = of rotate, fill=yellow-75] {
		\begin{tabular}{c}
		Horizontal Flip \\\hline
		$p=0.5$
		\end{tabular}
	};
	\node (imagewise) [above = of flip, yshift=-1cm] {Image-wise};
	
	\begin{scope}[on background layer]
	\node[base,dashed,fit = (pixelwise)(blur)(brightness)(hsv)(noise), minimum height=3cm]       (pixelwisebg) {};
	
	\node[base,dashed,fit = (imagewise)(flip)(rotate)(shift), minimum height=3cm]       (imagewisebg) {};
	\end{scope}
	
	\node (legend) [above = of imagewisebg, yshift=-0.75cm]{
		\begin{tabular}{|rp{3.8cm}|}
		\hline
		\multicolumn{2}{|l|}{\textbf{Legend}} \\\hline
		\cbox{black-25} & Disabled in AE \\
		\multicolumn{2}{|l|}{\textbf{Disabled in Categories:}}\\
		\cbox{yellow-75} & \textit{capsule}, \textit{metal nut}, \textit{pill}, \textit{screw} \\
		\cbox{rwth-50} & \textit{transistor} \\
		\cbox{magenta-50} & \textit{grid} \\
		\cbox{magenta} & \textit{bottle}, \textit{cable}, \textit{capsule}, \textit{metal nut}, \textit{pill}, \textit{screw}, \textit{toothbrush}, \textit{transistor}, \textit{wood}, \textit{zipper} \\\hline
		\end{tabular}
	};

	\draw[->] (blur) to (brightness);
	\draw[->] (brightness) to (hsv);
	\draw[->] (hsv) to (noise);
	\draw[->] (noise) to[out=0, in=180] (flip);
	\draw[->] (flip) to (rotate);
	\draw[->] (rotate) to (shift);

	\end{tikzpicture}
	\caption[Augmentation pipeline]{Augmentation pipeline used to enhance the MVTec AD dataset.
	Individual augmentation steps are applied consecutively, and $p$ denotes the probability of an augmentation being applied.
	Where multiple similar augmentations of the same type can be applied, columns indicate mutually exclusive choices, and the numbers inside a column their weights used during sampling.
	Augmentations are excluded for the categories where they induce an anomaly when applied (e.g. rotation for transistor).
	For further details, we refer to our implementation available at \url{https://github.com/ORippler/gaussian-ad-mvtec}}
	\label{fig:augmentation_pipeline}
\end{figure*}

\begin{table*}[htbp]
	\caption{AUROC scores in percent for all models per MVTec AD category.
	The AE approaches map-mean and CCA stand for the image-level aggregation by averaging or connected component analysis respecitvely.
	ResNet and EfficientNet are abbreviated as RN and EN.
	Mahalanobis approaches and oc-SVM are averaged over all levels to provide the simplest approach.
	The highest AUROC score per row is highlighted in bold.}
	\label{tab:final_results_categories}
	\centering
	\begin{tabular}{@{}llrrrrrrrrrr@{}}
		\toprule
		\multicolumn{2}{c}{\multirow{2}{*}{Score}}             & \multicolumn{2}{c}{MSE AE} & \multicolumn{4}{c}{\multirow{2}{*}{Pre-trained Classifier}} & \multicolumn{2}{c}{Mahalanobis (ours)} & \multicolumn{2}{c@{}}{{oc-SVM}} \\
		\cmidrule(lr){3-4}
		\cmidrule(l){9-10}
		&& Map-Mean & CCA & &&& &\multicolumn{1}{c}{All Features} & NPCA 1\% && \\
		\midrule
		\multicolumn{2}{r}{Architecture}        & RN-18           & RN-18      & RN-18              & RN-34 & EN-B0 & EN-B4 &  EN-B4 & EN-B4 &  EN-B0 & EN-B4               \\
		\midrule
		\parbox[t]{2mm}{\multirow{5}{*}{\rotatebox[origin=c]{90}{Textures}}}
		& Carpet     & 65.9                & 80.5           & 97.5                   & 96.8      & 97.2            & 98.9                         	& \textbf{100.0} & \textbf{100.0}	& 61.1	& 89.2 \\
		& Grid       & 80.9                & 92.9           & 90.2                   & 89.3      & 97.8            & \textbf{98.4}                	& 81.7           & 89.7				& 23.7	& 44.7 \\
		& Leather    & 46.0                & 90.5           & 99.8                   & 98.4      & 98.8            & 99.1                  		  	& 99.7           & \textbf{100.0}	& 75.8	& 86.7 \\
		& Tile       & 55.4                & 75.7           & 97.7                   & 99.1      & 99.2            & 97.8                   		& \textbf{99.8}  & \textbf{99.8}	& 91.6	& 95.7 \\
		& Wood       & 91.8                & 91.4           & 98.1                   & 95.0      & 95.6            & 99.6                         	& 98.6           & \textbf{99.6}	& 92.5	& 79.1 \\
		\midrule
		\parbox[t]{2mm}{\multirow{10}{*}{\rotatebox[origin=c]{90}{Objects}}}
		& Bottle     & 97.5                & 85.5           & 94.9                   & 98.2      & 99.5            & 99.5                  			& 99.8           & \textbf{100.0}	& 98.4	& 97.7 \\
		& Cable      & 79.5                & 58.3           & 91.1                   & 90.8      & 92.2            & 92.0                         	& \textbf{95.5}  & 95.0             & 78.3	& 81.0 \\
		& Capsule    & 74.5                & 76.7           & 92.4                   & 93.0      & 89.0            & \textbf{96.3}                  & 93.8           & 95.1             & 66.1	& 73.0 \\
		& Hazelnut   & 90.0                & 91.1           & 98.6                   & 98.8      & 98.4            & \textbf{99.8}                  & 99.6           & 99.1             & 83.5	& 81.7 \\
		& Metal Nut  & 58.2                & 64.6           & 95.6                   & 95.4      & 94.6			   & \textbf{96.8}                 	& 94.7   		 & 94.7             & 73.4	& 77.3 \\
		& Pill       & 80.1                & 58.0           & 86.0                   & 83.3      & 89.8            & \textbf{93.6}                	& 88.4           & 88.7             & 66.7	& 69.0 \\
		& Screw      & \textbf{95.7}       & 93.7           & 85.0                   & 88.8      & 90.3            & 95.2                         	& 85.4           & 85.2             & 19.5	& 31.0 \\
		& Toothbrush & 94.2                & \textbf{100.0} & 82.0                   & 86.9      & 80.7            & 85.4                			& 96.4           & 96.9             & 90.5	& 86.3 \\
		& Transistor & 85.9                & 79.1           & 92.4                   & 89.7      & 90.0            & 94.1                       	& \textbf{96.3}  & 95.5             & 82.0	& 83.8 \\
		& Zipper     & 86.8                & 88.8           & 97.9                   & \textbf{98.3}      & 98.8            & 97.9              	& 97.8           & 97.9             & 91.7	& 94.7 \\
		\midrule
		& Mean       & 78.8                & 81.8           & 93.3                   & 93.4      & 94.1            & \textbf{96.3}                	& 95.2           &  95.8	        & 73.0	& 78.1 \\
		& SEM        & 4.1                 & 3.4            & 1.4                    & 1.3       & 1.4             & 1.0                  			& 1.5            &  1.2             & 6.1	& 4.7  \\
		\bottomrule
	\end{tabular}
\end{table*}

\begin{figure*}[htbp]
	\setlength{\lineskip}{0pt}  
	\centering
	\includegraphics[height=0.3\textheight]{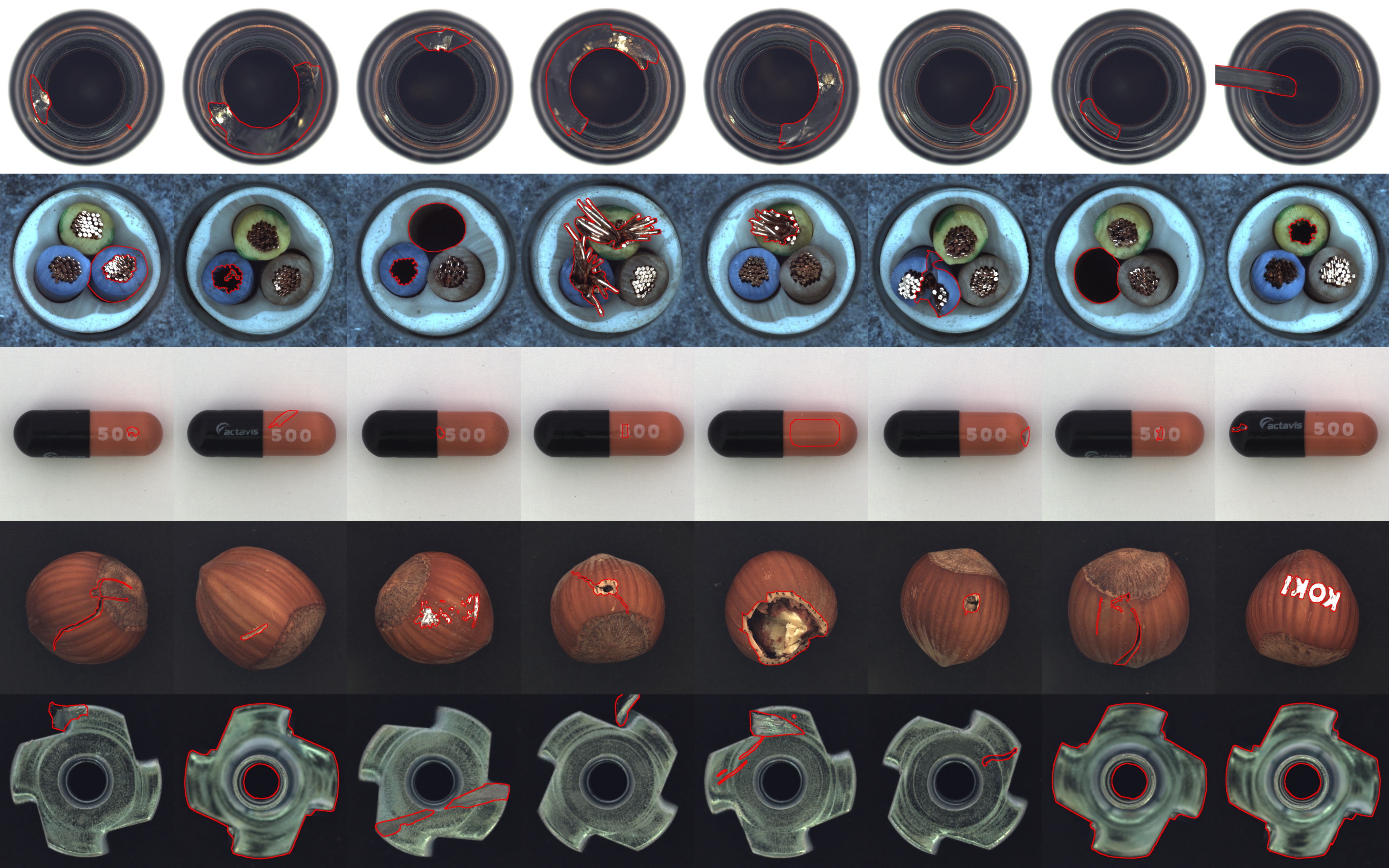}
	\includegraphics[height=0.3\textheight]{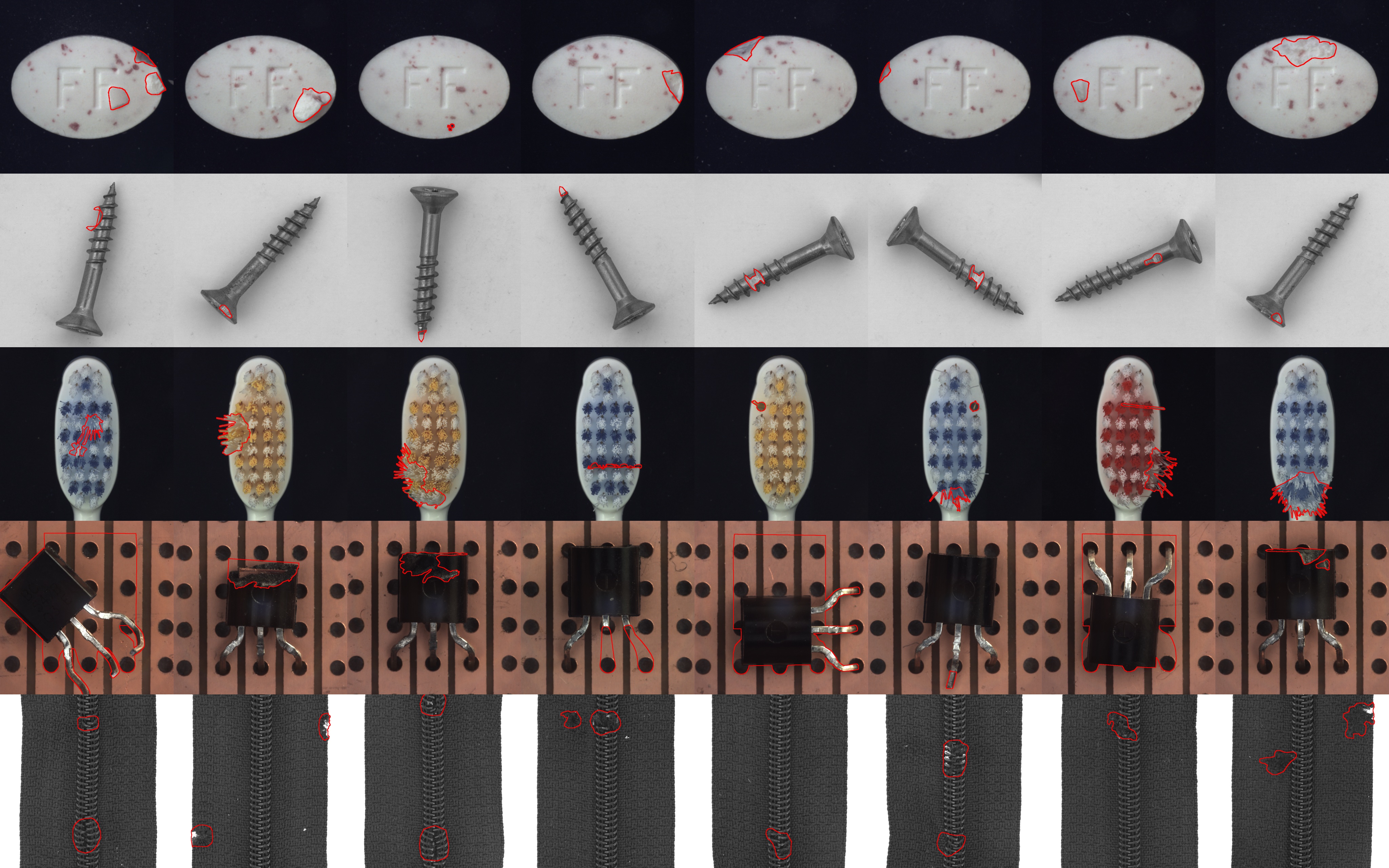}
	\includegraphics[height=0.3\textheight]{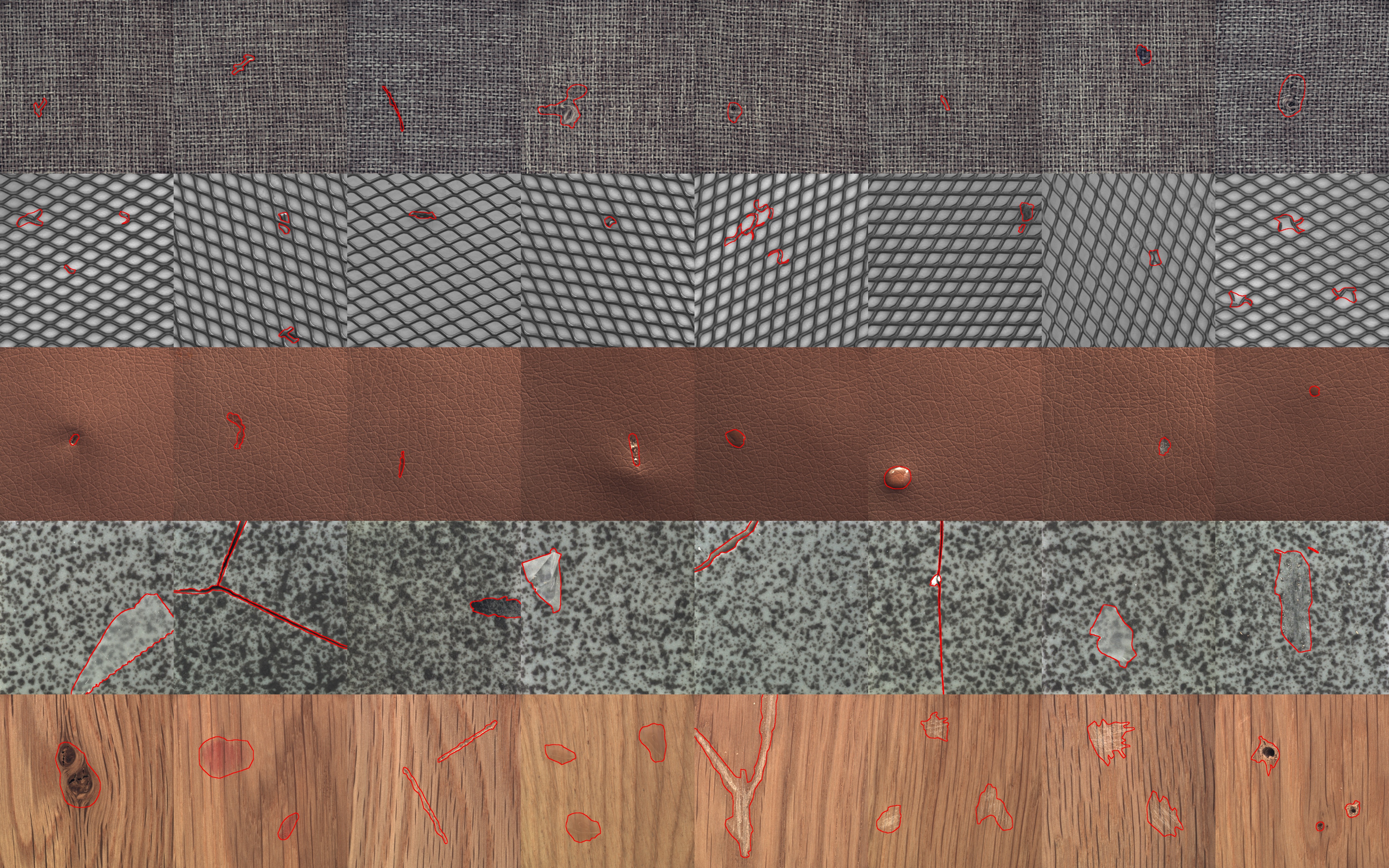}
	\caption[Augmentation pipeline]{Representative anomalies for all categories of the MVTec AD dataset. Red lines are boundaries of the segmentation ground truth. For additional information about the MVTec AD dataset, we refer to the original publication \cite{Bergmann2019}.}
	\label{fig:representative_anomalies}
\end{figure*}

\end{document}